\documentclass{article}

\usepackage{amsmath,amssymb,amsfonts}
\usepackage{algorithmic}
\usepackage{graphicx}
\usepackage{textcomp}
\usepackage{booktabs}
\usepackage{xcolor}
\usepackage{hyperref}
\usepackage{booktabs}
\usepackage{graphicx}
\usepackage{multirow}
\usepackage{amssymb}
\usepackage{makecell}

\usepackage{caption}
\usepackage{booktabs}
\usepackage{multirow}
\usepackage{graphicx}
\usepackage{subcaption}

\usepackage[preprint]{corl_2026} 

\title{Learning with Object-centric Representations of Tactile Interactive Perception for Robot Manipulation}

\author{
  Xinyi Yang$^{*1}$, Zilin Si$^{*1}$, Zhuowei Xu$^{1}$,
  Zeynep Temel$^{1}$, and Oliver Kroemer$^{1}$ \\
  $^{1}$Robotics Institute\\
  Carnegie Mellon University 
  United States\\
  \texttt{\{xinyiy4, zsi, zhuoweix, ztemel, okroemer\}@andrew.cmu.edu}
}

\begin{document}
\maketitle
{\let\thefootnote\relax\footnotetext{$^{*}$Equal contribution.}}
\begin{abstract}
Implicit object properties that are difficult to directly infer from vision, such as material, container contents, or softness, can be revealed through tactile sensing and exploratory interactions. However, because tactile signals are transient and sparse, extracting informative tactile events and effectively incorporating them into robotic manipulation remains a challenge.
In this work, we present an object-centric context-aware manipulation framework that learns task-agnostic object representations through tactile exploration. A token learner autonomously selects representative tactile segments from long-horizon exploration, while contrastive alignment with descriptive text embeddings enables a latent space that captures multiple physical object properties. These learned representations are then used as semantic context to guide object-centric manipulation policies and adapt strategies based on object properties.
Experiments show that the learned representations achieve 93\% and 84\% property estimation accuracy on seen and unseen objects. Evaluated on three tasks involving visually ambiguous objects, i.e. multi-object rearrangement, pouring, and box opening, the proposed framework improves both target selection and property-dependent manipulation adaptation, raising task success, aggregated over all evaluation trials, from 41\% to 92\% on seen objects and from 19\% to 67\% on unseen objects over baseline policies without object-context conditioning.
Videos and additional results are available at \url{https://xinyiyxyx.github.io/tactile-object-centric/}.
\end{abstract}



\section{Introduction}

Vision-based manipulation policies~\cite{chi2025diffusion, zhao2023learning, intelligence2025pi_} primarily rely on directly observable visual cues to predict robot actions. However, for visually ambiguous objects such as opaque or enclosed containers, vision alone is often insufficient, leading to uncertainty and degraded manipulation performance. In such scenarios, robots must rely on additional sensing modalities, such as tactile sensing, to infer hidden physical properties, including weight, stiffness, friction, and internal contents, that are inaccessible through vision alone~\cite{chen2016learning, huang2022understanding}. Access to such properties is crucial for selecting appropriate manipulation strategies, such as distinguishing empty from full containers for pouring or adapting grasping behaviors based on object rigidity. Prior work in interactive perception~\cite{katz2008manipulating, bohg2017interactive, xu2019densephysnet, tatiya2024mosaic} has shown that exploratory actions such as shaking, sliding, and tapping can reveal these object properties through physical interaction.

Despite the potential of tactile exploration, effectively exploiting tactile information remains challenging because tactile signals are sparse, transient, and event-driven~\cite{johansson2009coding}. Important object-related cues are therefore diluted in long-horizon exploration sequences, making compact and informative representation learning essential. 
Tactile-language alignment~\cite{yang2024bindingtoucheverythinglearning, cheng2025touch100k, ma2025cltp} has been shown to improve tactile representation learning by grounding embeddings in semantic physical properties. These language-aligned representations capture higher-level semantics beyond low-level sensory variations, improving transferability across interactions and downstream tasks.
Motivated by these, we propose a task-agnostic object-centric representation learning module that uses a learnable token-selection mechanism to extract representative tactile events and align the resulting object embeddings with semantic physical-property descriptors in a shared latent space.

In real-world manipulation settings, objects may share similar appearances, but require different actions depending on their physical properties and task intent. Therefore, it is crucial to incorporate the object context information into policy learning for decision-making, especially for multi-object manipulation~\cite{mitash2024scaling, wen2024object}.  We propose associating learned object-centric representations with task-relevant decisions to distinguish between ambiguous objects, select appropriate interaction strategies such as adjusting grasping and placement based on both object properties and task requirements, to improve the adaptability in multi-object manipulation tasks.

\begin{figure}
    \centering
    \includegraphics[width=0.99\linewidth]{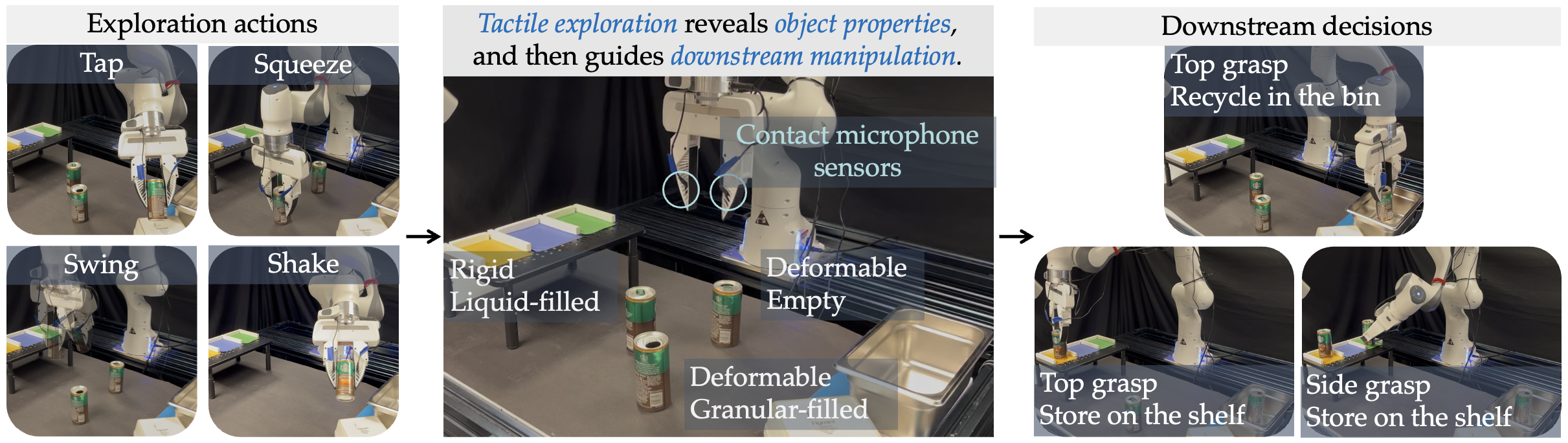}
    \caption{An object-centric manipulation learning framework. Task-agnostic object representations are learned through tactile exploration and then used as semantic context for downstream manipulation. In a multi-object rearrangement task, the policy adapts grasping strategies based on object rigidity and placement strategies based on object contents.}
    \label{fig:teaser}
    \vspace{-3mm}
\end{figure}

In this work, we present an object-centric manipulation framework that learns task-agnostic object representations from tactile exploration and incorporates them into a context-aware policy learning pipeline for downstream manipulation (Fig.~\ref{fig:teaser}). Our framework includes:
1) a learnable \textit{top-$K$} token-selection module that extracts compact, representative tactile contact-event tokens from long-horizon exploration for efficient object representation learning;
2) a contrastive alignment approach between tactile object embeddings and physical-property text embeddings, producing an interpretable semantic latent space that improves generalization across objects and interactions;
3) a context-aware policy learning model built on the Action Chunking Transformer~\cite{zhao2023learning}, integrating task conditioning and object association to enable adaptive manipulation under visual ambiguity.

We evaluate the framework in terms of both object property
prediction and downstream manipulation on three real-world
tasks: multi-object rearrangement and pouring with a
parallel-jaw gripper, and box opening with a multi-finger
robotic hand. Across eight training seeds, the learned
representations achieve mean property prediction accuracies
of 93\% and 84\% on seen and unseen objects.
For downstream manipulation, conditioning policies on
tactile-derived semantic context improves target selection
and property-dependent manipulation, raising task success from 41\% to 92\%
on seen objects and from 19\% to 67\% on unseen objects
compared with policies without object context.

\section{Related Work}

\textbf{Tactile Representation Learning}:
Self-supervised learning has been widely used to learn tactile representations across diverse sensing modalities, including acoustic sensing~\cite{thankaraj2023sounds}, vision-based tactile sensors~\cite{higuera2024sparsh}, and tactile sensing on robotic hands for dexterous manipulation~\cite{guzey2023dexterity, sharma2025self}. Inspired by representation learning methods in vision and multimodal learning~\cite{grill2020bootstrap, oquab2023dinov2}, these approaches pre-train tactile encoders on large-scale interaction data to capture contact-rich features that improve downstream policy learning. More recent work has explored language-aligned or semantically grounded tactile representations through contrastive learning, enabling tactile embeddings to be organized in more structured latent spaces and transferred across tasks~\cite{radford2021learning, yang2024bindingtoucheverythinglearning, cheng2025omnivtla, wu2025convitac}. While these methods primarily focus on learning representations from local contact observations or short-horizon interactions, we instead learn \textit{object-centric representations} from long-horizon exploratory interactions that can be flexibly used as semantic context in downstream manipulation.

\textbf{Tactile Perception and Object Property Inference}:
Beyond representation learning, tactile sensing has long been used to infer latent object properties such as material, compliance, weight distribution, and internal contents through exploratory interactions~\cite{sinapov2016learning, chen2016learning, huang2022understanding}. Earlier approaches often relied on Bayesian inference and probabilistic estimation for object identification and physical-property reasoning~\cite{xu2013tactile, eguiluz2018multimodal, pastor2020bayesian}. While these methods demonstrate the value of tactile exploration for recovering object attributes beyond visual perception, the inferred properties are typically used for recognition or state estimation rather than reusable object-level representations for downstream control. Since tactile exploratory trajectories are often long-horizon and sparse, compact representation learning becomes essential. Motivated by latent bottleneck architectures such as TokenLearner~\cite{ryoo2021tokenlearner} and Perceiver~\cite{jaegle2021perceiver}, we learn compact tactile tokens from exploration trajectories and align the resulting object embeddings with semantic physical-property descriptors.

\textbf{Robot Manipulation with Tactile Sensing}:
Tactile sensing has been increasingly integrated into learning-based robot manipulation to improve reactive control, multimodal policy learning, and contact-rich interaction~\cite{liu2024maniwav, zhao2025polytouch, chen2025implicitrdp, xue2025reactive,huang20253dvitaclearningfinegrainedmanipulation}. Prior work has fused tactile and visual observations using transformer-based architectures~\cite{zhao2025polytouch}, diffusion policies~\cite{liu2024maniwav}, and hierarchical or slow-fast control frameworks to improve reactive grasping, closed-loop adjustment, and robust contact-rich behaviors~\cite{chen2025implicitrdp, xue2025reactive, chen2026multimodalmanipulationmultimodalpolicy}. While these approaches primarily leverage tactile sensing as short-horizon feedback during task execution, they often do not explicitly incorporate tactile-derived object-level semantics into policy reasoning. Our work bridges tactile perception and manipulation by using object-centric tactile representations as semantic context for policy conditioning, while also studying the complementary role of online tactile observations for improving grasp quality during execution.

\section{Methods}

Our framework consists of two components: 1) A task-agnostic object representation learning module based on tactile interactive perception (Sec.\ref{sec:object-representation}). We aim to learn a latent space where each individual object and its multiple physical properties can be represented by a compact embedding vector in the space; and 2) a context-aware manipulation policy learner for acquiring manipulation behaviors that adapt based on the interactive perception information (Sec.\ref{sec:object-centric-policy}).
Here, task-agnostic means that the representation model is trained using physical-property supervision, without conditioning on downstream manipulation goals or actions.
\subsection{Object Representation Learning via Tactile Interactive Perception}
\label{sec:object-representation}

\begin{figure*}[h]
\centering
\includegraphics[width=0.99\linewidth]{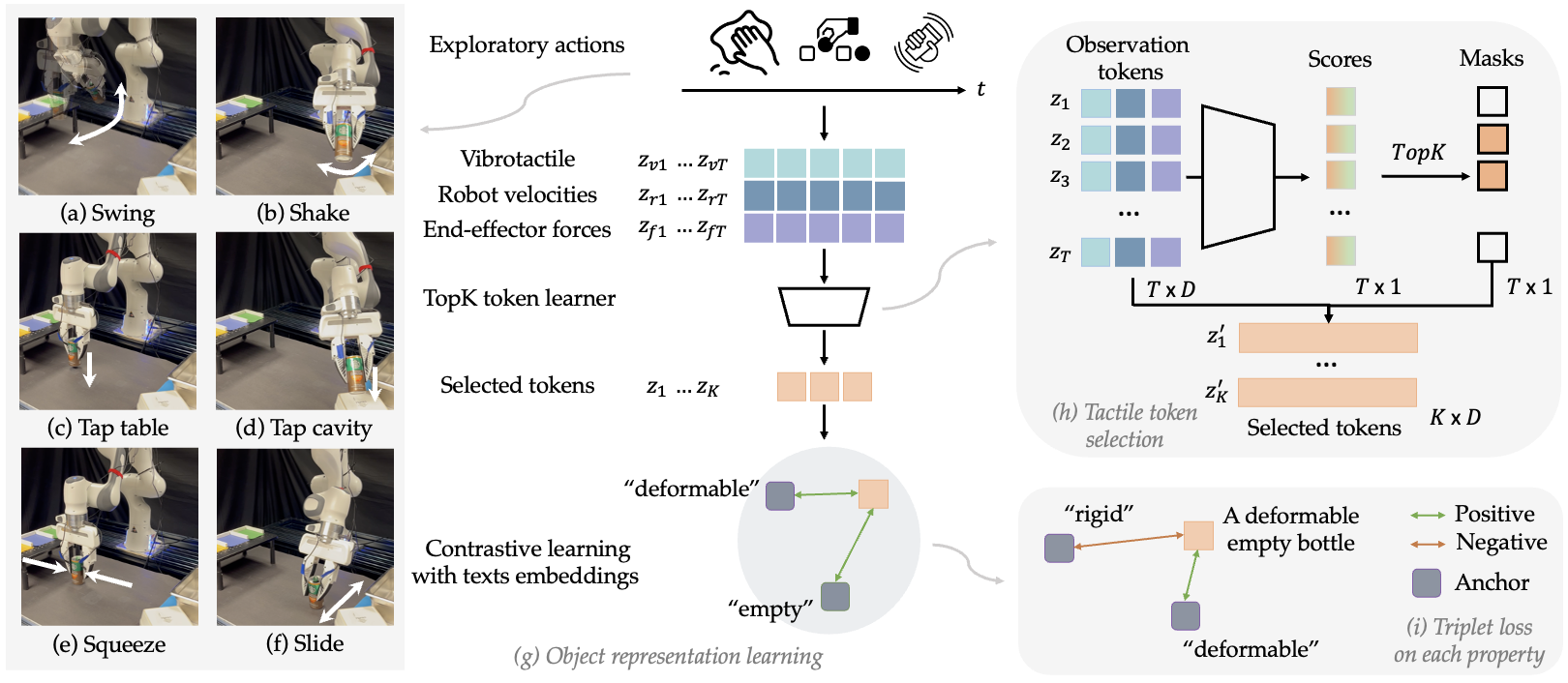}
\caption{Object representation learning through tactile exploration. Each object is explored with a sequence of pre-defined actions while vibrotactile signals, 6-axis end-effector force measurements, and robot end-effector linear velocities are recorded. A token learner selects the most informative $K$ tokens, which are aligned with text embeddings of physical-property descriptions using contrastive learning to produce semantic object representations.}
\label{fig:exploration-module}
\vspace{-5mm}
\end{figure*}

We aim to learn compact object embeddings that capture physical properties through exploratory tactile interaction. As shown in Fig.~\ref{fig:exploration-module}, each object is explored using a predefined action sequence while three synchronized modalities are recorded: (1) vibrotactile signals from contact microphones embedded in the gripper fingertips, (2) 6-axis end-effector force measurements derived from robot joint torques, and (3) robot end-effector linear velocities from the robot's joint encoders. Each modality is first encoded independently, then fused and compressed through a token learner to produce an object-level embedding.

\textbf{Object Exploration}: Each object is annotated along four property axes: \textit{content} (empty, solid, granular, liquid), \textit{fill level} (empty, half-filled, fully-filled), \textit{material} (plastic, metal, glass), and \textit{rigidity} (rigid, deformable). To probe these properties, we design six exploratory procedures (EPs) (Fig.~\ref{fig:exploration-module}(a-f)): (1) swinging, (2) shaking, (3) tabletop tapping, (4) cavity tapping on a ceramic surface to induce richer resonant responses, (5) squeezing, and (6) tabletop sliding. These motions are inspired by common human exploratory behaviors while remaining feasible for robotic execution.

\textbf{Vibrotactile, Force, and Robot Velocity Encoder}: The sequence of the exploration sensory readings is segmented into 0.5~s windows with 0.25~s overlap. For each segment, the raw vibrotactile data are converted to the frequency domain using STFT and encoded with a CNN backbone adapted from PANN~\cite{kong2020panns}, followed by projection into a latent token ($D_{vibro}=256$). Force and robot velocity readings are synchronized with the vibrotactile window and encoded using MLP modules to produce latent tokens with the same dimensions ($D_{force}=256$, $D_{velocity}=256$). The vibrotactile, force, and velocity tokens are then concatenated into a multimodal token sequence for token selection. More details of each encoder are in Appendix~\ref{sec:obj-represent-appendix}.

\textbf{Token Learner}: Exploratory interactions often generate long sensory measurements, while informative tactile events typically occur only at sparse contact transitions or dynamic interactions. Inspired by adaptive tokenization~\cite{ryoo2021tokenlearner}, we introduce a token learner (Fig.~\ref{fig:exploration-module}(h)) that selects $K$ most informative tokens from the exploration sequence to preserve these salient signals efficiently.
A learned scoring function assigns a relevance score to each token, and the $K$ highest-scoring tokens are selected in the forward pass.
To train the scoring network through this discrete selection, we use a straight-through estimator with a hard top-$K$ mask in the forward pass and a differentiable soft mask in the backward pass. The scoring MLP thus receives gradients for all tokens, including unselected ones, and is optimized by the triplet loss alone without any auxiliary selection loss.

The selected tokens are ordered by decreasing relevance score, concatenated, and projected into a compact object embedding ($D_{\text{obj}}=256$), followed by $\ell_2$ normalization for downstream use. We use $K=18$ in our experiments.

\textbf{Semantic Alignment with Text Embeddings}: To learn an interpretable latent space structured by language, we align object embeddings with text embeddings describing those properties (e.g., ``deformable'', ``empty'') by optimizing triplet loss on each property axis as shown in Fig.~\ref{fig:exploration-module}(i). Text embeddings are generated using a pre-trained off-the-shelf language model as anchors. 
To form triplets, within each batch across all property axes, we pair positive samples with matched properties and negative samples with mismatched properties with text anchors. All triplet losses from all property axes are aggregated for model training. During inference, object properties are estimated by computing the cosine similarity between object embeddings and candidate text embeddings.

\subsection{Object-Centric and Context-Aware Policy Learning}
\label{sec:object-centric-policy}

\begin{figure*}
\centering
\includegraphics[width=0.9\linewidth]{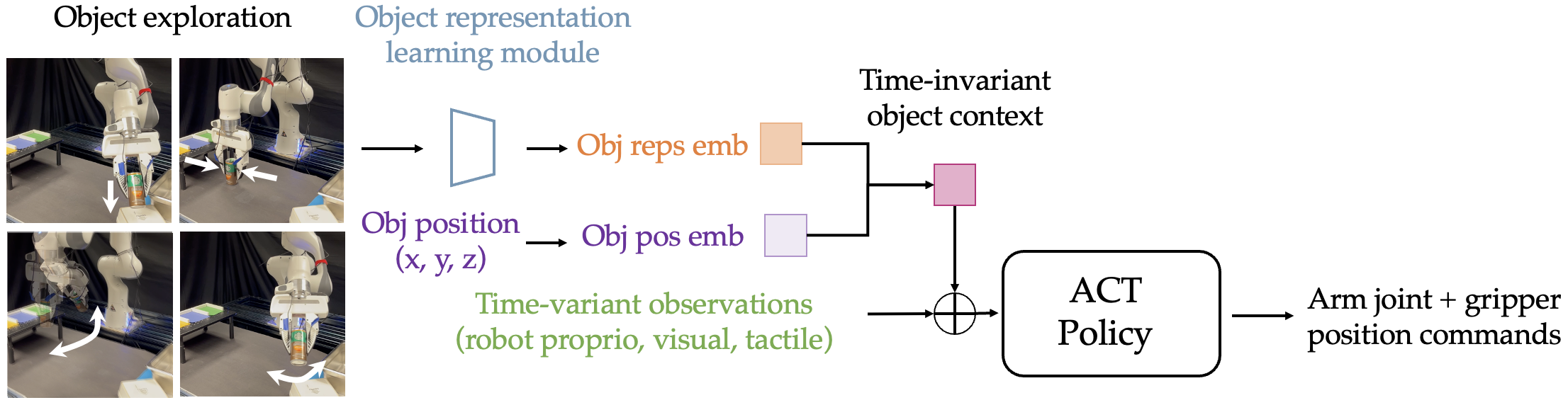}
\caption{Object-centric and context-aware manipulation policy learning. Built on the Action Chunking Transformer, the policy conditions on learned object context and object position to focus on target-specific visual, tactile, and proprioceptive observations for adaptive manipulation.}
\label{fig:policy-module}
\vspace{-5mm}
\end{figure*}

Manipulation becomes challenging when objects are visually ambiguous but require different actions based on physical properties. To address this, we incorporate learned object representations into a context-aware manipulation policy.

As shown in Fig.~\ref{fig:policy-module}, our policy model is built on the Action Chunking Transformer~\cite{zhao2023learning}, consisting of a transformer-based observation encoder and action decoder. To incorporate object-level context, we construct a time-invariant object context by concatenating the learned object representation with a positional embedding encoded from the object coordinates $(x,y,z)$. This object context conditions the policy jointly with time-varying visual, proprioceptive, and tactile observations. The policy then predicts action sequences consisting of 7D joint angles and a 1D gripper command.

In scenes containing multiple objects, we first explore each object to generate its corresponding representation. The policy then executes the task sequentially taking each object representation and its position as the object context. When the policy outputs idle actions, indicating that the current task has been completed, execution switches to the next object.

\section{Experiments}
\begin{figure*}[b]
\centering
\vspace{-5mm}
\includegraphics[width=1.0\linewidth]{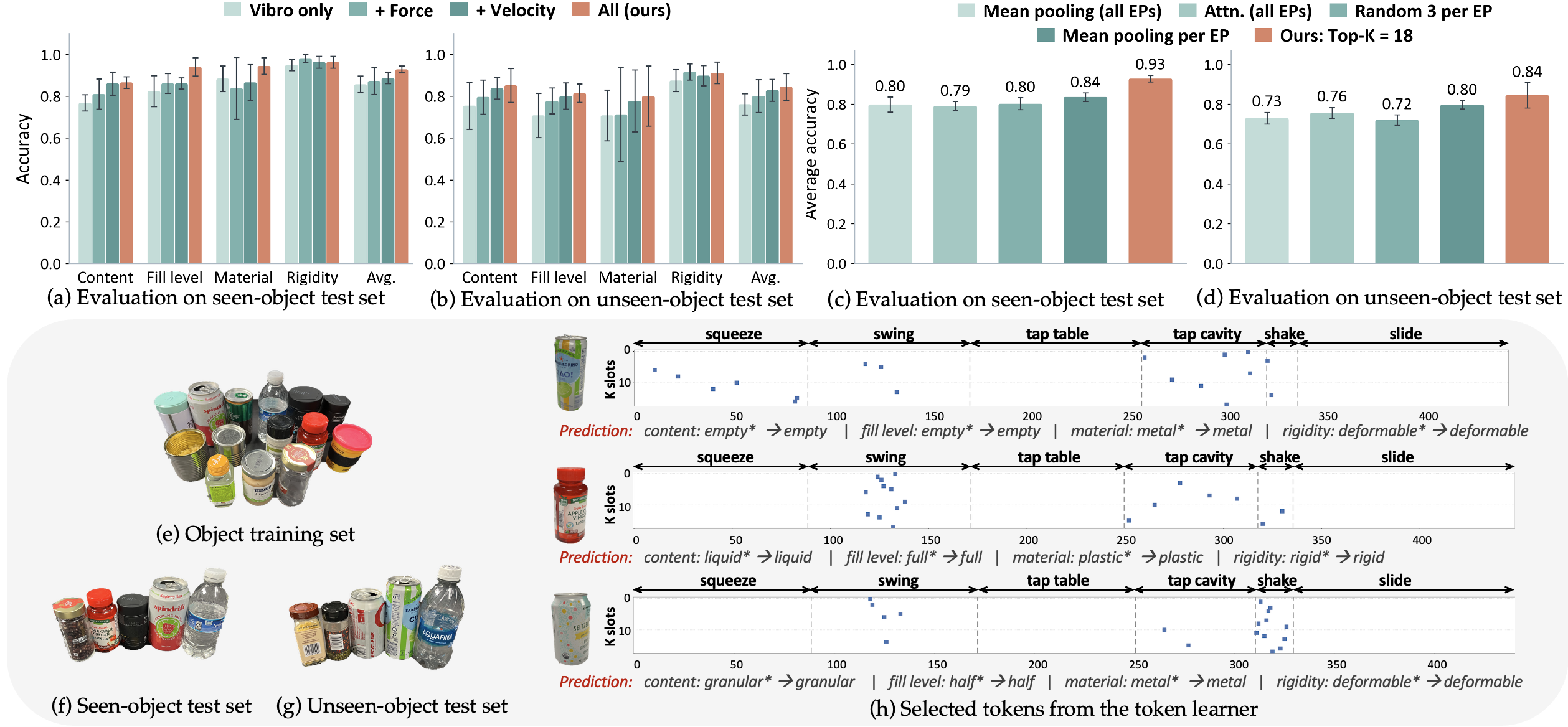}
\caption{Evaluation of learned tactile object representations. (a,b) Object property estimation on seen and unseen objects. Multi-modal sensing (vibrotactile, force, and velocity) achieves the best performance and generalizes to unseen objects. (c,d) Token learner ablation against trajectory-compression baselines. Bars in (a--d) show mean accuracy across eight training seeds (0--7), and error bars denote $\pm 1$ sample standard
deviation. Numerical results are provided in Table~\ref{tab:rep-ablation}. (e) Objects for model training. (f,g) Seen and unseen objects for evaluation. (h) Examples of selected representative tokens. This shows that the model prioritizes informative interactions such as squeezing, swinging, shaking, and cavity tapping.}
\label{fig:exploration-results}
\vspace{-5mm}
\end{figure*}
We evaluate the proposed framework in terms of both representation learning and downstream manipulation performance. Specifically, we study: (1) object property estimation accuracy to assess whether learned interactive tactile representations capture meaningful physical attributes and generalize to unseen objects; (2) token selection behavior to verify whether the token learner focuses on informative exploratory interactions; (3) manipulation performance and policy generalization on object rearrangement, pouring, and box opening tasks requiring property-conditioned target selection and adaptive manipulation strategies for visually ambiguous and unseen objects.

\subsection{Experimental Setup for Object Representation Learning}

\textbf{Robot Hardware Setup}:
Experiments are conducted on a real-world setup using a Franka Emika Panda arm equipped with a Franka Hand parallel-jaw gripper. The gripper uses fin-ray fingers, each embedded with a contact microphone sensor (Fig.~\ref{fig:teaser}) to capture vibrotactile signals. 

\textbf{Dataset and Data Collection}: We construct a tactile exploration dataset using everyday containers and household objects with diverse physical properties (Fig.~\ref{fig:exploration-results}(e)). In total, we collect 136 exploration trajectories across 23 object-property configurations. Random perturbations are introduced during motion generation to increase data diversity. Additional implementation details are provided in Appendix~\ref{sec:obj-represent-appendix}.

\subsection{Object Representation Learning Evaluation}
\label{sec:rep-eval}

\textbf{Object Property Estimation}:
We evaluate the learned representation precision by decoding object properties using cosine similarity between learned object embeddings and candidate text embeddings. The results are reported on five seen objects that are used for representation model learning (Fig.~\ref{fig:exploration-results}(f)) and five unseen hold-out objects (Fig.~\ref{fig:exploration-results}(g)). Each split comprises nine attribute configurations with
three evaluation trials per configuration, totaling 27 trials. We measure \textit{per-axis accuracy} across content, fill level, material, and rigidity across eight training seeds (0--7). As shown in Fig.~\ref{fig:exploration-results}(a,b), the learned representations achieve average accuracies of $0.93 \pm 0.02$ on seen objects and $0.84 \pm 0.06$ on unseen objects (mean $\pm$ std over 8 seeds), largely retaining the property estimation ability on novel objects.

\textbf{Modality Ablation}:
We compare different sensing combinations to understand the contribution of each modality. Using all three modalities of vibrotactile signals, forces, and end-effector velocities, yields improved performance compared to vibrotactile-only or pairwise combinations on both seen and unseen objects as shown in Fig.~\ref{fig:exploration-results}(a,b).
This suggests that force and velocity signals provide complementary context for interpreting tactile responses: vibrotactile sensing captures object reactions, while velocity and force describe excitation dynamics, contact state, and load variations.

\textbf{Token Selection Visualization}:
To determine which interactions provide the most informative context tokens, we map selected tokens back to their corresponding exploration segments (Fig.~\ref{fig:exploration-results}(h)). 
Rather than attending uniformly across the exploration trajectory, the results show that the token learner prioritizes short, information-dense events from squeezing, swinging, shaking, and cavity tapping, preserving physically meaningful tactile cues.
Furthermore, in downstream manipulation tasks, the learned token selection enables the use of only a subset of exploration actions, reducing exploration effort while preserving informative tactile cues.

\textbf{Token Selection Ablation}:
To isolate the benefits of learnable token selection, we compare the proposed \textit{top-K} token learner with four compression baselines: (1) mean pooling over all EPs, (2) attention over all EPs, (3) random selection of three tokens per EP, and (4) mean pooling within each EP. As shown in Fig.~\ref{fig:exploration-results}(c,d), the proposed method achieves the highest overall accuracy on both seen and unseen objects. These results indicate that performance gains come from selecting informative tactile events rather than simple trajectory compression.

\textbf{Comparison with Supervised Classification}:
To assess the effect of text alignment beyond categorical
property supervision, we train a cross-entropy (CE) baseline
that shares our encoder and token learner but differs in
the output head and training objective
(Appendix~\ref{sec:ce-appendix}).
Across eight training seeds, mean accuracy averaged over
the four property axes is comparable ($0.92$/$0.85$ for CE versus $0.93$/$0.84$ for ours,
seen/unseen). 

\begin{figure}[t]
    \centering
    \includegraphics[width=0.99\linewidth]{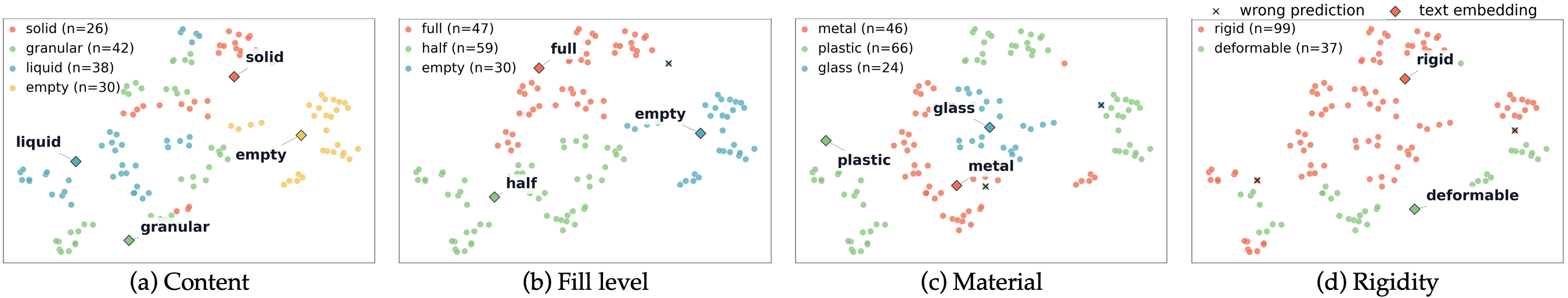}
    \caption{Visualization of learned object embeddings and aligned text embeddings on each property axis using UMAP dimension reduction. Embedding clusters show semantic alignment between tactile representations and text anchors.}
    \label{fig:embeddings}
    \vspace{-5mm}
\end{figure}

\textbf{Embedding Space Visualization}:
We visualize the learned object embedding space using UMAP~\cite{mcinnes2018umap}, jointly projecting object embeddings and text embeddings into the 2D space. As shown in Fig.~\ref{fig:embeddings}, the learned embedding space exhibits meaningful clustering by physical property, with object embeddings distributed close to corresponding semantic text anchors.

\textbf{Evaluation with a Multi-Finger Robotic Hand}: We evaluate the same framework with a multi-finger robotic DeltaHand~\cite{si2024deltahands} equipped with tactile sensors on the fingertips~\cite{xu2025multi} to show that the proposed framework can be extended to dexterous embodiments with in-hand manipulation exploration. Overall, we achieve 96\% accuracy for seen objects and 88\% for unseen objects on object property estimation. More details on data collection and results can be seen in the Appendix~\ref{sec:obj-represent-hand-appendix}.

\subsection{Experimental Setup for Manipulation Tasks}

\begin{figure*}[t]
\centering
\includegraphics[width=0.99\linewidth]{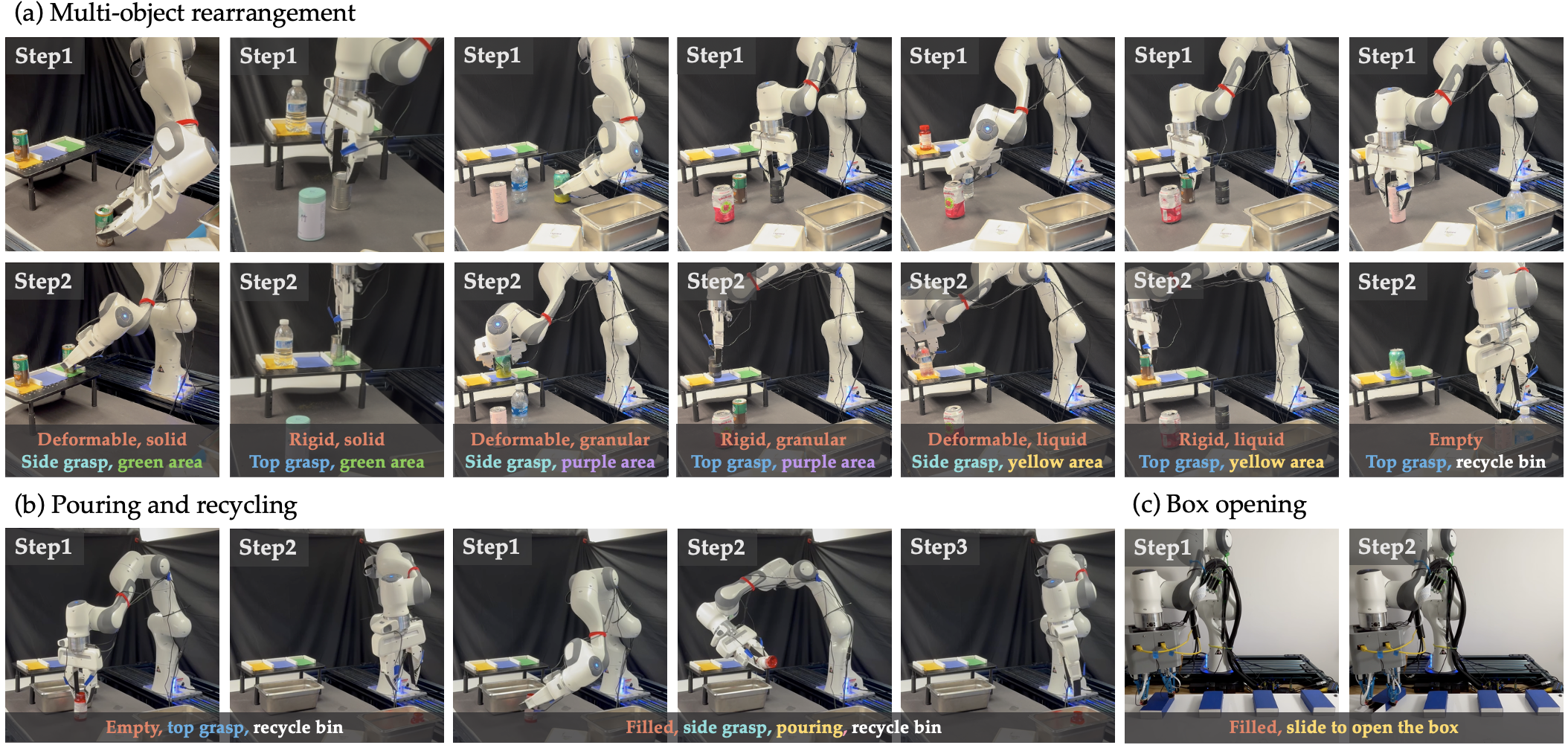} 
\caption{(a) Multi-object rearrangement task. The policy selects appropriate grasping strategies and places objects into task-specific target regions based on their rigidity and contents.
(b) Pouring task. The policy chooses a top grasp for the bottle if it is empty and directly recycles it, or uses a side grasp to pour out the liquid before recycling it.
(c) Box opening task with a multi-finger hand. The policy chooses to open heavy boxes containing screws and takes no action for empty boxes.}
\label{fig:task}
\vspace{-3mm}
\end{figure*}

\textbf{Task Setup}:
We evaluate the policy learning on three manipulation tasks: multi-object rearrangement and pouring with a parallel-jaw gripper, and box opening with a multi-finger hand~\cite{xu2025multi}. In the multi-object rearrangement task (Fig.~\ref{fig:task}(a)), three objects are placed on the table, which may share similar or different appearances and may contain the same or different internal contents. Manipulation decisions depend jointly on rigidity and the objects' internal content, producing seven possible grasp-and-placement strategies. For the pouring task (Fig.~\ref{fig:task}(b)), a single object is placed on the table, and the robot must either pour-then-recycle or directly recycle  the bottle depending on the bottle's fill level. For the box opening task (Fig.~\ref{fig:task}(c)), given four visually identical boxes, the hand approaches each box, then slides the box open if it is filled, or takes no action if it is empty.

\textbf{Data Collection}:
We collect demonstrations using a Factr~\cite{liu2025factr} teleoperation device for seen objects in multi-object rearrangement and pouring tasks (Fig.~\ref{fig:policy-results}(a) and (c)), resulting in 180 and 80 demonstrations respectively. For the box opening task, the robot arm reaching motions are generated through a planner and the in-hand motions are teleoperated through a kinematic-twin device~\cite{si2024tilde}. In total, 50 demonstrations are collected. More policy training details can be seen in Appendix~\ref{sec:policy-appendix}.

\subsection{Manipulation Policy Learning Evaluation}
\label{sec:policy-eval}
\textbf{Policy Evaluation Metrics and Ablation}:
We evaluate policy performance using two metrics: \textit{decision accuracy} measures correct strategy selection (grasp type and goal location) given object properties, and \textit{task success rate} measures end-to-end success, requiring both correct decisions and successful execution without dropping or misplacing the object. 

To analyze the effects of semantic abstraction, representation quality, and reactive tactile sensing, we compare six policy configurations:
(1) \textit{Baseline w/o context}, using only target object positions;
(2) \textit{Baseline w/o obj position}, using only object context, i.e., the tactile-interaction tokens;
(3) \textit{w/ tactile context, vision only}, using learned tactile embeddings as object context;
(4) \textit{w/ text context, vision only}, using decoded semantic text embeddings from tactile embeddings via nearest-neighbor cosine similarity;
(5) \textit{w/ tactile/text context, vision + vibrotactile}, adding online vibrotactile observations as reactive inputs; and
(6) \textit{Oracle w/ GT text context}, using ground-truth text embeddings to isolate downstream execution performance from representation errors.
To evaluate generalization, we conduct rollouts on both training objects and ``unseen'' objects. Here \emph{unseen} is defined with respect to policy training, i.e., novel object instances or property configurations absent from the demonstrations; see Appendix~\ref{sec:task-appendix} for details.
\begin{figure}[t]
    \centering
    \includegraphics[width=0.98\linewidth]{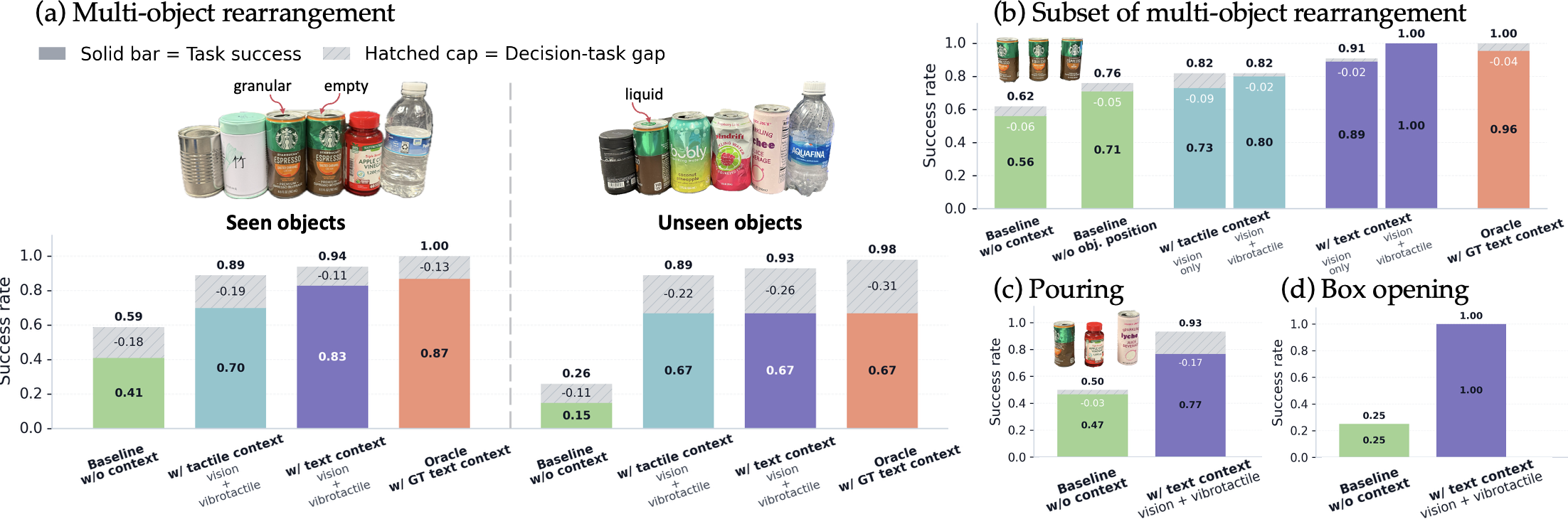}
    \caption{Policy evaluation results. (a) Performance of the multi-object rearrangement task with seen and unseen objects. (b) Ablation with a subset of visually identical objects on context type and reactive tactile sensing. (c, d) Performance of the pouring and box opening tasks. }
    \label{fig:policy-results}
    \vspace{-5mm}
\end{figure}

\textbf{Main Results}: Results are shown in Fig.~\ref{fig:policy-results}. We achieve improved task success rates using semantic object contexts ($0.83$ (seen) and $0.67$ (unseen), $0.77$, and $1.00$), compared to baselines without context ($0.41$ (seen) and $0.15$ (unseen), $0.47$, and $0.25$) for multi-object rearrangement, pouring, and box opening tasks. Aggregated over all evaluation trials in Table~\ref{tab:policy-results-all}, the text-context policy (vision + tactile) succeeds in 154/167 seen-object trials ($0.92$) and 43/64 unseen-object trials ($0.67$), versus 69/167 ($0.41$) and 12/64 ($0.19$) for the baseline without context. Additional task descriptions and result analysis can be seen in Appendix~\ref{sec:task-appendix}.

\textbf{Text-Aligned Object Contexts Improve Policy Stability}:
As shown in Fig.~\ref{fig:policy-results}(a, b), using text-aligned semantic context consistently improves policy success over directly conditioning on raw tactile embeddings ( (a) $0.70 \rightarrow 0.83$, $0.67 \rightarrow 0.67$, (b) $0.73 \rightarrow 0.89$, $0.80 \rightarrow 1.00$). Due to the high-dimensional latent space, tactile embeddings of objects with the same physical properties can still vary because of shape differences, exploration noise, or execution variability. With limited demonstrations, this can create out-of-distribution conditions during inference. By contrast, projecting tactile representations onto semantic text anchors provides a many-to-one mapping, reducing latent variation and yielding more consistent, largely symbolic decision-making that is less sensitive to minor context variations, thereby improving policy conditioning and generalization.

\textbf{Text-Aligned Context vs. Labeled Context}:
To assess whether the policy gains stem from semantic alignment or merely from conditioning on any labeled context, 
we compare policies conditioned on one-hot versus text context in multi-object rearrangement
(Table~\ref{tab:policy-results-set1}). At inference time, the one-hot context is predicted by CE baseline described in Sec.~\ref{sec:rep-eval}, encoded as one-hot vectors. One-hot context is competitive in distribution (0.91/0.78 vs. 0.94/0.83 strategy/task success on seen objects) but brittle on unseen objects (0.80/0.48 vs. 0.93/0.67). Notably, one-hot context fails more often on unseen objects even when the correct context is retrieved, typically through imprecise edge grasps. 
One possible explanation is that the policy overfits to associations between one-hot property combinations and object-specific grasps, which transfer poorly to unseen geometries. Language-aligned context may mitigate this effect by capturing semantic relationships among property
values, supporting transfer across object configurations. However, whether this representation encourages greater reliance on visual geometry for grasp placement remains
to be tested.

\textbf{Reactive Tactile Sensing Improves Task Execution}: Comparing policies with identical object context but different observation inputs (\textit{vision only} vs.\ \textit{vision + tactile}) in Fig.~\ref{fig:policy-results}(b), we observe consistent gains in success rates ($0.73 \rightarrow 0.80$, $0.89 \rightarrow 1.00$). Vibrotactile feedback provides immediate contact information that is difficult to infer from vision alone. In failure cases of vision-only policies, the gripper often failed to close firmly enough, leading to unstable grasps and slipping.

\textbf{Generalization to Unseen Objects through Symbolic Decisions}:
From evaluation on unseen objects with novel shapes and physical configurations using a policy trained on seen objects (Fig.~\ref{fig:policy-results}(a)), we observe that decision accuracy remains high ($0.93$ vs.\ $0.98$), and task success rates are identical between predicted semantic context and oracle context ($0.67$ vs.\ $0.67$). Although task success decreases compared to seen objects, most failures come from low-level execution rather than semantic misclassification. Future improvements may come from broader demonstration diversity and geometry-aware grasp refinement.

\section{Conclusions}
We presented an object-centric manipulation framework that combines tactile exploration-based representation learning with context-aware policy learning for visually ambiguous manipulation tasks. Through tactile exploration, the proposed method learns compact object representations that capture meaningful physical properties. Experimental results show that the token learner effectively identifies informative tactile signals from long-horizon exploration, enabling accurate object property estimation and generalization to unseen objects. 
Furthermore, using these learned object representations as semantic context conditions allows the downstream policy to select targets and adapt manipulation strategies based on object properties in visually ambiguous tasks. 

\textbf{Limitations}:
The current framework relies on predefined exploration procedures and offline data collection, with a separate exploration phase that adds time before manipulation. 
The robot does not yet select exploratory procedures based on task requirements or request additional observations when its object context is insufficient for a decision.
Future work will investigate task-driven online exploration that acquires information as needed during manipulation, together with policy learning that supports more flexible
semantic conditioning. In addition, scaling policy learning to larger multi-object scenarios requires more demonstrations, motivating sample-efficient approaches such as synthetic data generation from limited demonstrations~\cite{xue2025demogen}.

\clearpage

\bibliography{reference}  

\clearpage
\appendix
\section*{Appendix}
\section{Object Representation Learning: Implementation Details}
\label{sec:obj-represent-appendix}

\subsection{Exploration Data Collection}
\paragraph{Exploration Data Augmentation}

During object exploration data collection, we apply motion augmentation as shown in Table~\ref{tab:random-parameter} to improve the generalization of the learned representations.

\begin{table*}[ht]
    \centering
    \caption{Data randomization configurations for object exploration data collection}
    \label{tab:random-parameter}
    \resizebox{\textwidth}{!}{
    \begin{tabular}{lcccccc}
        \toprule
        Parameter 
        & Swing 
        & Shake 
        & Tap on cavity 
        & Tap on table 
        & Squeeze 
        & Slide \\
        \midrule
        
        Amplitude 
        & $[20,28]^\circ$ 
        & $[20,30]^\circ$ 
        & -- 
        & -- 
        & -- 
        & -- \\
        
        Lifting distance 
        & -- 
        & -- 
        & $[3,3.3]$ cm 
        & $[1.5,2.5]$ cm 
        & -- 
        & -- \\
        
        Grasp force 
        & -- 
        & -- 
        & -- 
        & -- 
        & 0.1 N 
        & -- \\
        
        Sliding distance 
        & -- 
        & -- 
        & -- 
        & -- 
        & -- 
        & $[5,8]$ cm \\
        
        Period 
        & $[0.8,1.0]$ s 
        & $[0.35,0.5]$ s 
        & $[0.2,0.8]$ s 
        & $[0.4,0.6]$ s 
        & -- 
        & $[1.6,2.2]$ s \\
        
        Repetitions 
        & 3 
        & $[3,5]$ 
        & $[5,8]$ 
        & $[4,7]$ 
        & 3 
        & 3 \\
        
        \bottomrule
    \end{tabular}
    }
\end{table*}

\paragraph{Data Configuration Details}
To collect object exploration data, we use 14 objects, shown in Fig.~\ref{fig:exploration-results}(e). Each object has a fixed material; we vary its content, fill level, and rigidity by leaving it empty or filling it with substances such as water, pasta, and beans. In total, the training pool covers 23 distinct attribute configurations (content, fill level, material, rigidity) over 136 exploration trials (Table~\ref{tab:train_configs}). Pairing the same container with different contents and fill levels ensures that a given container appears under multiple attribute configurations, which discourages the encoder from taking a shortcut based on container identity alone. For representation learning, we use an 80/20 train/validation split stratified by attribute configuration.

To evaluate the learned representation, we independently collect two exploration test sets (Fig.~\ref{fig:exploration-results}(f) and (g)): a \emph{Seen} set using five objects from the training pool, and an \emph{Unseen} set using five novel containers never used during training. Both sets cover the same nine attribute configurations with three trials each, constructing 27 trials per set (Table~\ref{tab:eval_configs}). Per-axis label distributions of the training pool and the evaluation sets are summarized in Table~\ref{tab:axis_dist_all}.

\begin{table}[htbp]
  \centering
  \caption{Per-axis label distribution of the training pool and the evaluation sets. The \emph{Seen} and \emph{Unseen} sets share identical distributions and are shown in a single row.}
  \label{tab:axis_dist_all}
  \setlength{\tabcolsep}{3.5pt}
  \small
  \begin{tabular}{lccccccccccccc}
    \toprule
    & \multicolumn{4}{c}{Content} & \multicolumn{3}{c}{Fill level} & \multicolumn{3}{c}{Material} & \multicolumn{2}{c}{Rigidity} & \\
    \cmidrule(lr){2-5} \cmidrule(lr){6-8} \cmidrule(lr){9-11} \cmidrule(lr){12-13}
    Split & empty & gran. & liquid & solid & empty & half & full & glass & metal & plastic & deform. & rigid & Total \\
    \midrule
    Train          & 30 & 42 & 38 & 26 & 30 & 59 & 47 & 24 & 46 & 66 & 37 & 99 & 136 \\
    Seen / Unseen  & 12 &  9 &  6 &  0 & 12 &  9 &  6 &  6 &  6 & 15 & 12 & 15 &  27 \\
    \bottomrule
  \end{tabular}
\end{table}
\begin{table}[htbp]
    \centering
    \caption{Attribute configurations of the object exploration data used for representation learning and evaluation.}
    \label{tab:exploration-data-configs}

    \begin{subtable}[t]{0.49\textwidth}
        \centering
        \caption{Training pool (136 trials over 23 configurations). Configurations marked with $\dagger$ also appear in the evaluation sets.}
        \label{tab:train_configs}
        \scriptsize
        \setlength{\tabcolsep}{4pt}
        \renewcommand{\arraystretch}{1.10}
        \begin{tabular}{llllc}
            \toprule
            \textbf{Content} & \textbf{Fill level} & \textbf{Material} & \textbf{Rigidity} & \textbf{Trials} \\
            \midrule
            empty    & empty & glass   & rigid      & 5$^\dagger$ \\
            empty    & empty & metal   & deformable & 5$^\dagger$ \\
            empty    & empty & plastic & deformable & 8$^\dagger$ \\
            empty    & empty & plastic & rigid      & 12$^\dagger$ \\
            granular & full  & glass   & rigid      & 4$^\dagger$ \\
            granular & full  & metal   & rigid      & 5 \\
            granular & full  & plastic & rigid      & 5 \\
            granular & half  & glass   & rigid      & 5 \\
            granular & half  & metal   & deformable & 7$^\dagger$ \\
            granular & half  & metal   & rigid      & 4 \\
            granular & half  & plastic & deformable & 4 \\
            granular & half  & plastic & rigid      & 8$^\dagger$ \\
            liquid   & full  & metal   & rigid      & 6 \\
            liquid   & full  & plastic & rigid      & 5$^\dagger$ \\
            liquid   & half  & glass   & rigid      & 5 \\
            liquid   & half  & metal   & deformable & 5 \\
            liquid   & half  & metal   & rigid      & 5 \\
            liquid   & half  & plastic & deformable & 4$^\dagger$ \\
            liquid   & half  & plastic & rigid      & 8 \\
            solid    & full  & glass   & rigid      & 5 \\
            solid    & full  & metal   & rigid      & 5 \\
            solid    & full  & plastic & rigid      & 12 \\
            solid    & half  & metal   & deformable & 4 \\
            \midrule
            \multicolumn{4}{l}{\textbf{Total}} & \textbf{136} \\
            \bottomrule
        \end{tabular}
    \end{subtable}
    \hfill
    \begin{subtable}[t]{0.49\textwidth}
        \centering
        \caption{Evaluation sets. Both sets share the same nine configurations with three trials each (one trial per placement location). \emph{Seen} uses training objects at new locations; \emph{Unseen} uses novel physical objects with the same attribute configurations.}
        \label{tab:eval_configs}
        \scriptsize
        \setlength{\tabcolsep}{4pt}
        \renewcommand{\arraystretch}{1.10}
        \begin{tabular}{llllcc}
            \toprule
            \textbf{Content} & \textbf{Fill level} & \textbf{Material} & \textbf{Rigidity} & \textbf{Seen} & \textbf{Unseen} \\
            \midrule
            empty    & empty & glass   & rigid      & 3 & 3 \\
            empty    & empty & metal   & deformable & 3 & 3 \\
            empty    & empty & plastic & deformable & 3 & 3 \\
            empty    & empty & plastic & rigid      & 3 & 3 \\
            granular & full  & glass   & rigid      & 3 & 3 \\
            granular & half  & metal   & deformable & 3 & 3 \\
            granular & half  & plastic & rigid      & 3 & 3 \\
            liquid   & full  & plastic & rigid      & 3 & 3 \\
            liquid   & half  & plastic & deformable & 3 & 3 \\
            \midrule
            \multicolumn{4}{l}{\textbf{Total}} & \textbf{27} & \textbf{27} \\
            \bottomrule
        \end{tabular}
    \end{subtable}
\end{table}
\subsection{Model Architecture}
The model takes a sequence of 0.5~s window of vibrotactile readings ($D_{raw}=22050$ at a 44.1~kHz sampling rate) with 0.25~s overlap. Signals are converted to the frequency domain using STFT (320 hop size, 64 mel bins). During training, we apply data augmentation by randomly dropping time segments and spectral stripes. The resulting spectrograms are encoded using a CNN backbone adapted from PANN~\cite{kong2020panns}, followed by projection into vibrotactile tokens ($D_{vibro}=256$).

End-effector force and velocity signals are sampled at 100~Hz, producing 50 samples per window. Force consists of 6 channels (3D force and 3D torque), and velocity consists of 3D linear velocity. Both modalities are synchronized with each vibrotactile frame using the frame end timestamp via ROS and encoded using MLP encoders (force: 6--256--256, velocity: 3--256--256). Gaussian noise ($\sigma=0.02$) is added to normalized force and velocity inputs. The vibrotactile, force, and velocity tokens are concatenated into a unified multimodal token sequence for token selection.

The fused token sequence is projected to a hidden dimension of $D_{hidden}=256$, and an MLP-based scoring function (256--1) assigns an importance score to each token. The top-$K=18$ tokens are selected, flattened, concatenated, and projected into a compact object embedding ($D_{obj}=256$), followed by $\ell_2$ normalization for downstream tasks.

We train the model using triplet loss to align object embeddings with text embeddings representing physical properties (e.g., “deformable”, “empty”). Text embeddings are obtained using a pre-trained SentenceTransformer (all-MiniLM-L6-v2). For each property axis, within-batch positives and negatives are constructed based on matching or mismatching labels, and all axis-wise losses are aggregated. A margin of 0.2 is used for the triplet loss. During inference, object properties are estimated via cosine similarity between object embeddings and candidate text embeddings. 

\subsection{Object Representation Training and Evaluation Protocol}
We use an 80/20 train-validation split stratified by object properties. The model is trained for 200 epochs with a batch size of 32. We use the Adam optimizer with a learning rate of $1\times10^{-4}$ and zero weight decay, along with a CosineAnnealingLR learning rate scheduler.

\section{Object Representation Learning: Additional Experiments}

\subsection{Open-Vocabulary Property Decoding} 
To test whether the text-aligned embedding space supports queries beyond the training vocabulary, we replace the decoding vocabulary with 29 paraphrases of the property descriptions (Table~\ref{tab:paraphrase-vocab}). 
Averaged over eight training seeds, accuracy drops modestly from 0.93/0.84 (seen/unseen) with the training prompts to 0.88/0.81 with paraphrases, and to 0.85/0.79 on the 17-paraphrase subset that shares no word stem with any training prompt (e.g., ``steel'' for ``metal''; marked in italics in Table~\ref{tab:paraphrase-vocab}). This suggests that text alignment places tactile embeddings in the semantic space of the language model rather than binding them to specific symbols, enabling open-vocabulary property queries that a categorical classifier cannot express without retraining.

\vspace{-3mm}
\begin{table}[htbp]
\centering
\caption{Training prompts and the 29-paraphrase inference vocabulary. The \emph{empty} class of the content and fill-level axes shares one training prompt; its paraphrases are split between the two axes during evaluation. Paraphrases in \emph{italics} share no word stem with any training prompt and form the strictest evaluation subset (17 in total).}
\label{tab:paraphrase-vocab}
\small
\setlength{\tabcolsep}{5pt}
\begin{tabular}{@{}llp{0.22\textwidth}p{0.44\textwidth}@{}}
\toprule
\textbf{Axis} & \textbf{Class} & \textbf{Training prompts} & \textbf{Inference paraphrases} \\
\midrule
\multirow{3}{*}{content}
  & solid    & solid content      & a solid object inside; one solid piece inside \\
  & granular & granular; particle & \emph{grains inside}; \emph{powder}; \emph{pellets} \\
  & liquid   & liquid; fluid      & \emph{water inside}; sloshing liquid \\
\cmidrule(lr){1-4}
content / fill level
  & empty    & empty              & \emph{hollow}; \emph{nothing inside}; \emph{contains nothing}; \emph{no contents}; \emph{vacant}; unfilled \\
\cmidrule(lr){1-4}
\multirow{2}{*}{fill level}
  & full     & full; fully filled & filled to the top; completely filled \\
  & half     & half               & at half capacity; \emph{midway level} \\
\cmidrule(lr){1-4}
\multirow{3}{*}{material}
  & metal    & metal              & metallic; made of metal; \emph{steel} \\
  & plastic  & plastic            & made of plastic; \emph{polymer} \\
  & glass    & glass              & made of glass; glassy \\
\cmidrule(lr){1-4}
\multirow{2}{*}{rigidity}
  & rigid    & rigid              & \emph{stiff}; \emph{unbending}; \emph{firm} \\
  & deformable & deformable       & \emph{squishy}; \emph{pliable} \\
\bottomrule
\end{tabular}
\end{table}
\vspace{-3mm}

\begin{table*}[t]
\centering
\caption{Object property estimation ablations on seen (S) and unseen (U)
objects at different locations. Values are mean $\pm$ sample standard
deviation over 8 seeds (0--7). Avg. is the mean accuracy across the four
property axes; its standard deviation is computed across the per-seed averages.}
\label{tab:rep-ablation}
\small
\setlength{\tabcolsep}{5pt}
\begin{tabular}{@{}lccccc@{}}
\toprule
\textbf{Method (split)}
& \textbf{Content}
& \textbf{Fill Level}
& \textbf{Material}
& \textbf{Rigidity}
& \textbf{Avg.} \\
\midrule

\multicolumn{6}{@{}l}{\textit{Modality ablation}} \\
Vibrotactile only (S)
& $0.77 \pm 0.04$ & $0.82 \pm 0.07$ & $0.88 \pm 0.06$
& $0.95 \pm 0.03$ & $0.86 \pm 0.04$ \\
Vibrotactile only (U)
& $0.75 \pm 0.11$ & $0.71 \pm 0.11$ & $0.71 \pm 0.12$
& $0.88 \pm 0.05$ & $0.76 \pm 0.05$ \\
\addlinespace[2pt]
Vibrotactile + force (S)
& $0.81 \pm 0.07$ & $0.86 \pm 0.05$ & $0.84 \pm 0.15$
& $0.98 \pm 0.02$ & $0.87 \pm 0.06$ \\
Vibrotactile + force (U)
& $0.80 \pm 0.08$ & $0.78 \pm 0.06$ & $0.71 \pm 0.23$
& $0.92 \pm 0.04$ & $0.80 \pm 0.08$ \\
\addlinespace[2pt]
Vibrotactile + velocity (S)
& $0.86 \pm 0.06$ & $0.86 \pm 0.03$ & $0.87 \pm 0.09$
& $0.96 \pm 0.03$ & $0.89 \pm 0.03$ \\
Vibrotactile + velocity (U)
& $0.84 \pm 0.05$ & $0.80 \pm 0.06$ & $0.78 \pm 0.15$
& $0.90 \pm 0.05$ & $0.83 \pm 0.05$ \\

\midrule
\multicolumn{6}{@{}l}{\textit{Token selection ablation}} \\
Mean pooling over all EPs (S)
& $0.76 \pm 0.09$ & $0.76 \pm 0.06$ & $0.76 \pm 0.05$
& $0.92 \pm 0.02$ & $0.80 \pm 0.04$ \\
Mean pooling over all EPs (U)
& $0.69 \pm 0.06$ & $0.69 \pm 0.04$ & $0.64 \pm 0.05$
& $0.90 \pm 0.03$ & $0.73 \pm 0.03$ \\
\addlinespace[2pt]
Attention over all EPs (S)
& $0.77 \pm 0.05$ & $0.71 \pm 0.03$ & $0.75 \pm 0.04$
& $0.93 \pm 0.03$ & $0.79 \pm 0.02$ \\
Attention over all EPs (U)
& $0.72 \pm 0.04$ & $0.68 \pm 0.06$ & $0.69 \pm 0.06$
& $0.95 \pm 0.03$ & $0.76 \pm 0.03$ \\
\addlinespace[2pt]
Random 3 tokens per EP (S)
& $0.73 \pm 0.05$ & $0.78 \pm 0.05$ & $0.77 \pm 0.11$
& $0.93 \pm 0.02$ & $0.80 \pm 0.03$ \\
Random 3 tokens per EP (U)
& $0.73 \pm 0.07$ & $0.74 \pm 0.05$ & $0.51 \pm 0.06$
& $0.91 \pm 0.04$ & $0.72 \pm 0.03$ \\
\addlinespace[2pt]
Mean pooling per EP (S)
& $0.80 \pm 0.03$ & $0.85 \pm 0.03$ & $0.81 \pm 0.06$
& $0.88 \pm 0.03$ & $0.84 \pm 0.02$ \\
Mean pooling per EP (U)
& $0.81 \pm 0.06$ & $0.79 \pm 0.04$ & $0.68 \pm 0.06$
& $0.90 \pm 0.03$ & $0.80 \pm 0.02$ \\

\midrule
\multicolumn{6}{@{}l}{\textit{Full model}} \\
\textbf{Ours} (all modalities, top-$K$) (S)
& $0.87 \pm 0.03$ & $0.94 \pm 0.04$ & $0.94 \pm 0.04$
& $0.96 \pm 0.03$ & $0.93 \pm 0.02$ \\
\textbf{Ours} (all modalities, top-$K$) (U)
& $0.85 \pm 0.08$ & $0.81 \pm 0.04$ & $0.80 \pm 0.14$
& $0.91 \pm 0.05$ & $0.84 \pm 0.06$ \\
\bottomrule
\end{tabular}
\end{table*}

\subsection{Comparison with Supervised Classification}
\label{sec:ce-appendix}
\begin{figure}[h]
  \centering
  \includegraphics[width=0.9\linewidth]{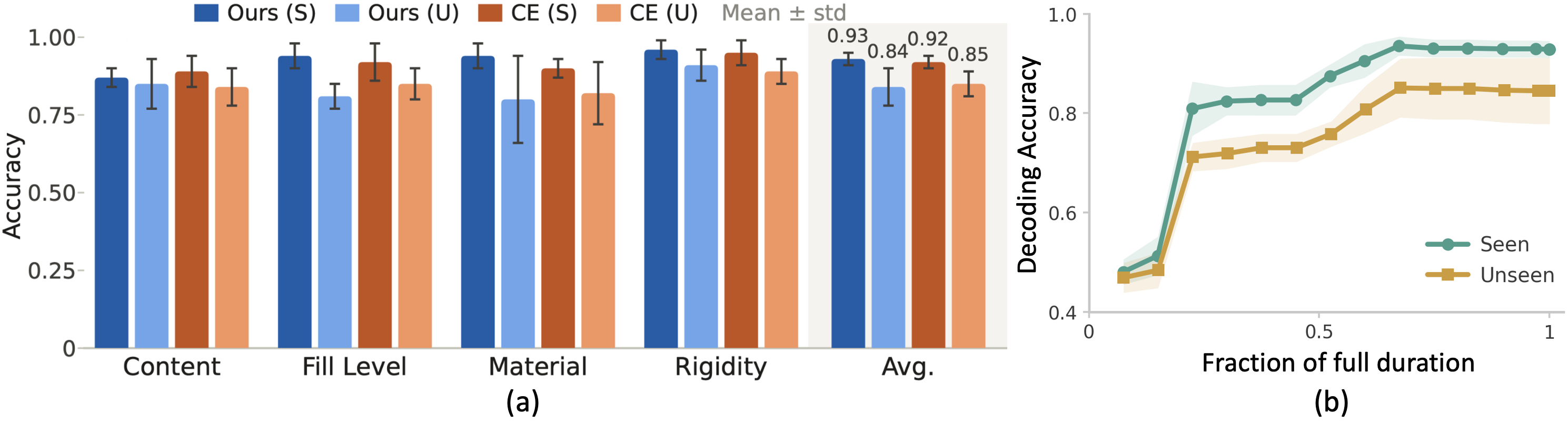}
  \caption{(a) Per-axis property estimation accuracy of ours and the CE baseline on seen (S) and unseen (U) objects, mean $\pm$ std over 8 seeds. (b) Axis-average accuracy vs.\ exploration time with a causal
top-$K$ token buffer decoded every 15\,s and at trajectory completion,
using only observation windows completed by the decoding time.
Accuracy is averaged over four object attributes and 27 test
trajectories per split. Curves show the mean across eight training
seeds; shaded bands indicate $\pm 1$ standard deviation across seeds.
The horizontal axis expresses the decoding times as a fraction of
full duration, computed as the mean elapsed exploration time divided
by the mean full trajectory duration within each split.}
  \label{fig:ce-comparison-exploration-tradeoff}
  \vspace{-3mm}
\end{figure}

To isolate whether our gains come from semantic alignment or merely from labeled supervision, we build a supervised classifier baseline that shares the encoder, token learner, training data, and hyperparameters with our model and differs only in the output head and objective. For the CE baseline, the four per-axis linear heads are trained with cross-entropy (CE) on one-hot labels and decoded by argmax, instead of projection into the text embedding space trained with the triplet loss and decoded by cosine similarity to text anchors. Both models are retrained with seeds 0--7 and evaluated on the same seen and unseen sets. Average accuracy is comparable ($0.92 \pm 0.02$ / $0.85 \pm 0.04$ for CE vs.\ $0.93 \pm 0.02$ / $0.84 \pm 0.06$ for ours on seen and unseen test set, respectively, as shown in Fig.~\ref{fig:ce-comparison-exploration-tradeoff}(a) and Table~\ref{tab:multiseed}). For both methods, the variance concentrates on the material axis of the unseen split, where a single glass object drives most errors. However, the failure modes differ qualitatively: the CE baseline makes errors on the glass object in all 8 seeds (6/24 correct, wrong-class average softmax 0.73), i.e., a systematic and confident error, whereas ours has no failure persisting across seeds (worst object: 15/24) and its errors are near-tied anchors (cosine deficit $<0.2$), i.e., boundary jitter rather than systematic shift. 

\begin{table}[h]
  \centering
  \caption{Numerical values of Fig.~\ref{fig:ce-comparison-exploration-tradeoff}(a).}
  \label{tab:multiseed}
  \small
  \setlength{\tabcolsep}{4pt}
  \begin{tabular}{@{}lccccc@{}}
    \toprule
    Method & Content & Fill Level & Material & Rigidity & Avg. \\
    \midrule
    Ours (Seen)   & $0.87 \pm 0.03$ & $0.94 \pm 0.04$ & $0.94 \pm 0.04$ & $0.96 \pm 0.03$ & $0.93 \pm 0.02$ \\
    Ours (Unseen) & $0.85 \pm 0.08$ & $0.81 \pm 0.04$ & $0.80 \pm 0.14$ & $0.91 \pm 0.05$ & $0.84 \pm 0.06$ \\
    CE (Seen)     & $0.89 \pm 0.05$ & $0.92 \pm 0.06$ & $0.90 \pm 0.03$ & $0.95 \pm 0.04$ & $0.92 \pm 0.02$ \\
    CE (Unseen)   & $0.84 \pm 0.06$ & $0.85 \pm 0.05$ & $0.82 \pm 0.10$ & $0.89 \pm 0.04$ & $0.85 \pm 0.04$ \\
    \bottomrule
  \end{tabular}
\end{table}

\subsection{Exploration Time--Accuracy Trade-off}
Full exploration with six exploratory procedures takes
$223 \pm 6$\,s per object. Token-selection statistics on the
training split indicate that tapping on the table and sliding are
largely redundant for the four studied property axes; omitting
them at deployment reduces exploration to $138 \pm 4$\,s without
retraining, while preserving property prediction accuracy.
To quantify the accuracy--cost trade-off more finely, we replay
test-set trajectories while maintaining a causal top-$K$ token
buffer and decode object properties every 15\,s, as shown in
Fig.~\ref{fig:ce-comparison-exploration-tradeoff}(b).

\section{Object Representation Learning Extension to a Multi-Finger Hand}
\label{sec:obj-represent-hand-appendix}
\begin{figure*}
\centering
\includegraphics[width=0.95\linewidth]{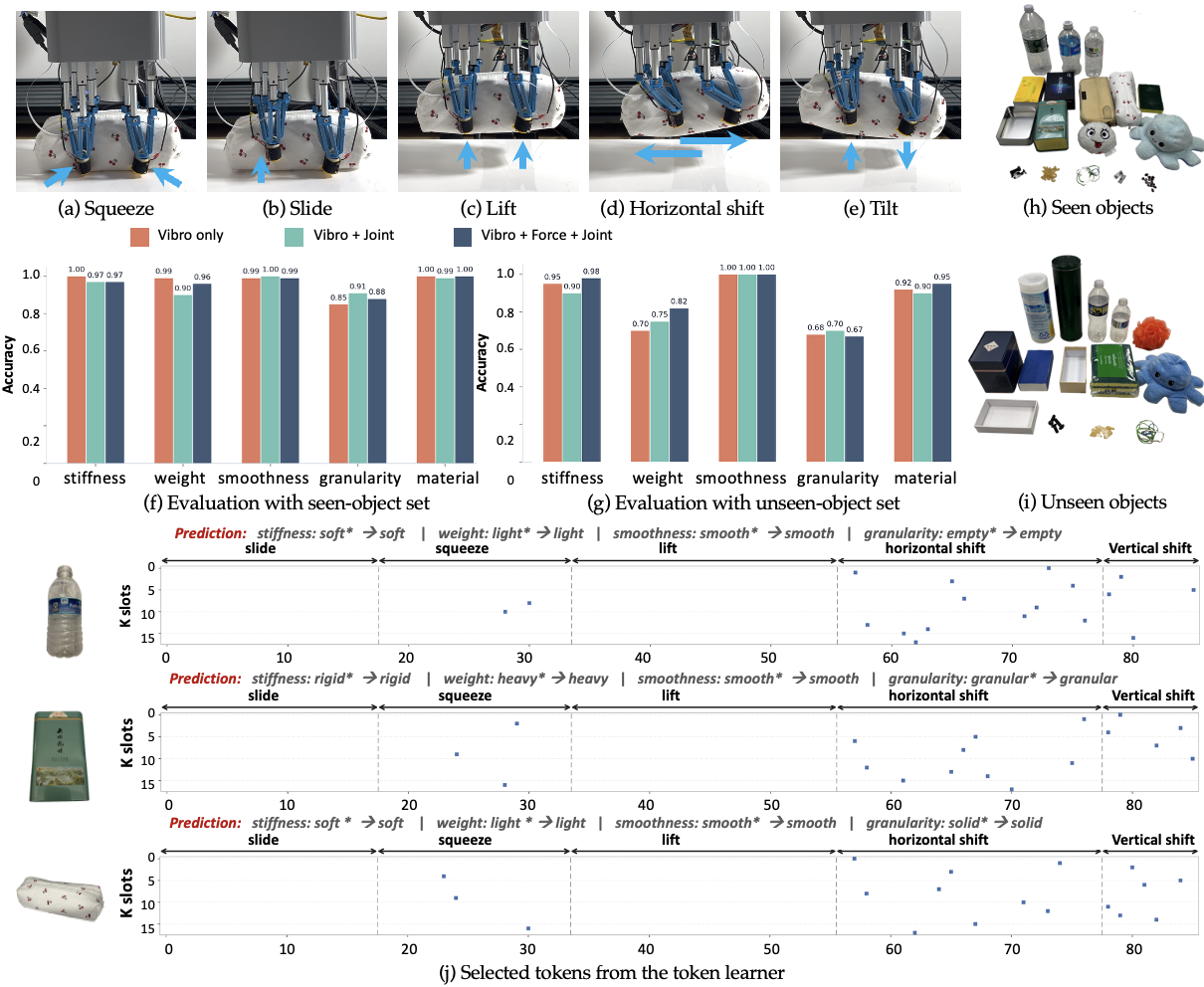}
\caption{Object representation learning with a multi-finger robotic hand (DeltaHand~\cite{si2024deltahands}). (a–e) Five exploration procedures: squeezing the object by closing all fingers, sliding one finger along the object surface while the remaining fingers provide support, lifting the object, in-hand horizontal shifting, and in-hand tilting. (f, g) Object property estimation is evaluated on both (h) seen and (i) unseen objects. (j) Three examples of selected tokens from the token learner are visualized, showing that most tokens are selected from squeezing, horizontal shifting, and tilting motions.}
\label{fig:deltahand-obj-repres-results}
\vspace{-9mm}
\end{figure*}

\textbf{Robot Hardware Setup}:
Experiments are conducted on a real-world setup using a Franka Emika Panda arm mounted with a DeltaHand~\cite{si2024deltahands}. The DeltaHand has four fingers, each equipped with a contact microphone and strain gauges to capture both dynamic contact events and static 3D force measurements~\cite{xu2025multi}. We record fingertip vibrotactile signals, force measurements, and finger joint positions from motor encoders.

\textbf{Object Exploration}:
Each object is annotated along five property axes: \textit{stiffness} (rigid, soft), \textit{weight} (light, heavy), \textit{smoothness} (smooth, rough), \textit{granularity} (empty, granular, solid), and \textit{material} (plastic, metal, paper, fabric). We collect 135 exploration trajectories over 27 object-property configurations, with 5 trials per configuration. The training dataset is shown in Fig.~\ref{fig:deltahand-obj-repres-results}(h).

To probe these properties, we design five exploratory procedures (EPs) (Fig.~\ref{fig:deltahand-obj-repres-results}(a–e)): (1) squeezing, (2) sliding, (3) lifting, (4) in-hand horizontal shifting, and (5) in-hand tilting (vertical shift). In contrast to parallel-jaw gripper motions that often rely on external environmental interactions (e.g., tabletop sliding or cavity tapping), we introduce in-hand exploration strategies such as horizontal shifting and tilting, enabling more efficient object interaction.

\textbf{Object Property Estimation}:
We evaluate representation quality by predicting object properties via cosine similarity between learned object embeddings and candidate text embeddings. Results are reported on 12 seen objects (Fig.~\ref{fig:deltahand-obj-repres-results}(h)) and 11 unseen objects (Fig.~\ref{fig:deltahand-obj-repres-results}(i)), with three evaluation trials per object. We report per-axis accuracy across properties. As shown in Fig.~\ref{fig:deltahand-obj-repres-results}(f,g), the model achieves 96\% accuracy on seen objects and 88\% on unseen objects on average. It generalizes well to stiffness, smoothness, and material, but shows reduced performance on weight and granularity.

\textbf{Modality Ablation}:
We evaluate different sensing combinations to analyze the contribution of each modality. While comparable performance is observed using vibrotactile-only, force+joint, or vibrotactile+joint inputs on seen objects (Fig.~\ref{fig:deltahand-obj-repres-results}(f)), incorporating all three modalities yields the best performance on unseen objects (Fig.~\ref{fig:deltahand-obj-repres-results}(g)).

\textbf{Token Selection Visualization}:
We map selected tokens back to their corresponding exploration segments (Fig.~\ref{fig:deltahand-obj-repres-results}(j)). Tokens are predominantly selected from squeezing, horizontal shifting, and tilting motions, indicating that in-hand interactions produce highly informative tactile events. This further suggests that robotic hands enable more efficient object exploration by reducing reliance on external environmental interactions.

\section{Manipulation Policy Learning}
\subsection{Implementation Details}
\label{sec:policy-appendix}

\begin{figure*}[h]
\centering
\includegraphics[width=0.85\linewidth]{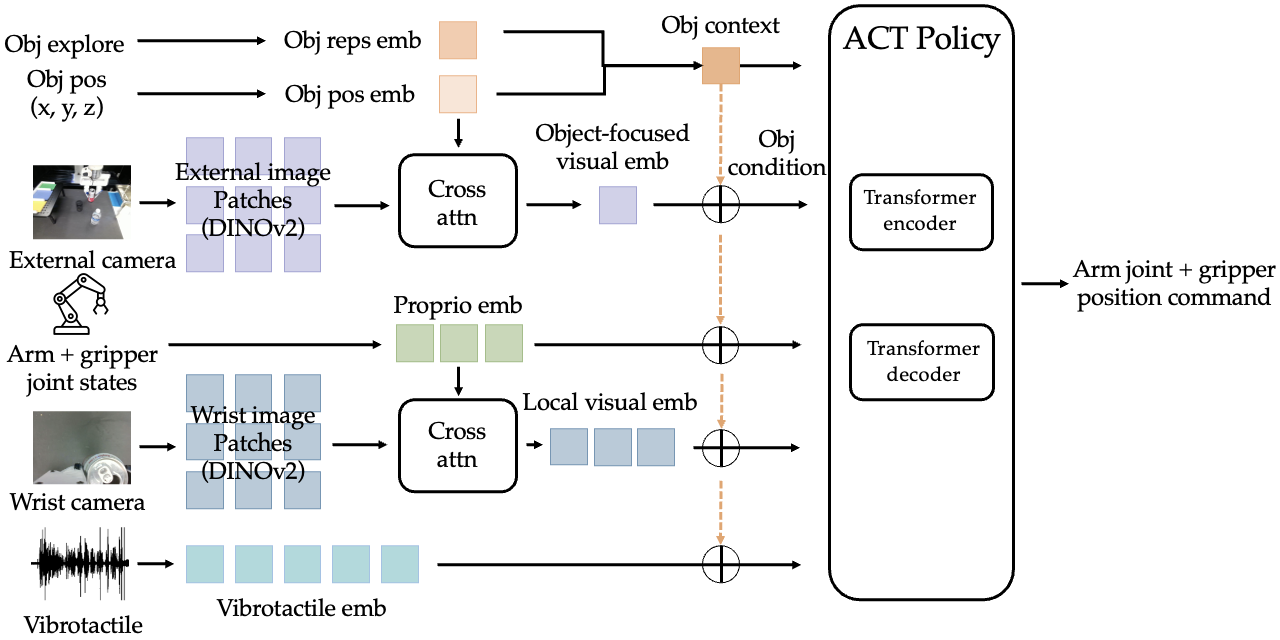}
\caption{Object-centric and context-aware manipulation policy learning. Built on the Action Chunking Transformer, the policy conditions on learned object context and object position to focus on target-specific visual, tactile, and proprioceptive observations for adaptive manipulation.}
\label{fig:policy-module-detail}
\vspace{-3mm}
\end{figure*}

\paragraph{Policy Architecture} A detailed policy architecture is shown in Fig.~\ref{fig:policy-module-detail}. To incorporate object-level context, we construct a time-invariant object context by concatenating the learned object representation with a positional embedding encoded from object coordinates $(x,y,z)$. This context conditions the policy alongside time-varying visual, proprioceptive, and tactile observations in a FiLM-style manner.

To improve target focus and balance cross-modal token interactions, we use the object position embedding to attend over external image tokens, producing an object-centric visual embedding. In addition, proprioceptive features attend to wrist-camera tokens to extract local visual features (8 tokens in our implementation).

For reactive tactile inputs, we process a 1.0~s window of vibrotactile signals, convert them to the frequency domain using STFT, and encode them using a CNN backbone adapted from PANN~\cite{kong2020panns}, followed by projection into a latent token ($D_{vibro}=256$).

The policy predicts 7D arm joint actions and a 1D gripper width in action chunks of size 64.

\paragraph{Policy Training}
Policies are trained with a batch size of 32 using an 80/20 train-validation split for 50 epochs. We use the Adam optimizer with a learning rate of $4\times10^{-4}$ and weight decay of $1\times10^{-4}$.

All inputs and outputs are normalized to $[-1,1]$, except images, which are normalized to $[0,1]$. During training, we apply data augmentation, including color jitter, random rotation, and Gaussian noise on images; Gaussian noise on robot proprioception and object locations; and spectral augmentations with stripe dropping for vibrotactile inputs.

For text-conditioned policies, we obtain object property embeddings from the corresponding text descriptions and aggregate them via mean pooling into a single embedding. For tactile-conditioned policies, we randomly sample precomputed object embeddings from the representation learning dataset with matching physical properties, eliminating the need for additional object exploration data during policy training.

\paragraph{Policy Evaluation Protocol}
To evaluate manipulation policies, we first collect object exploration trajectories using only the actions selected by the token learner. We then condition the policy on either inferred tactile embeddings or their corresponding text embeddings. During inference, object context is fully inferred from deployment exploration unless marked oracle. The policy predicts action chunks of length 64, while 32 actions are executed at 15 Hz. Results are reported in Table~\ref{tab:policy-results-all}.

\begin{table*}
    \centering
    \caption{Manipulation policy evaluation across different task settings.}
    \label{tab:policy-results-all}

    \begin{subtable}[t]{0.49\textwidth}
        \centering
        \caption{Multi-object rearrangement task.}
        \label{tab:policy-results-set1}
        \scriptsize
        \setlength{\tabcolsep}{3.2pt}
        \renewcommand{\arraystretch}{1.10}
        \begin{tabular}{lcccc}
            \toprule
            \multirow{2}{*}{\textbf{Method}} &
            \multicolumn{2}{c}{\textbf{Strategy}} &
            \multicolumn{2}{c}{\textbf{Overall}} \\
            \cmidrule(lr){2-3}
            \cmidrule(lr){4-5}
            & Count & Rate & Count & Rate \\
            \midrule
            \multicolumn{5}{l}{\textbf{Seen objects} (6 objects)} \\
            \midrule
            Baseline w/o context & 32/54 & 0.59 & 22/54 & 0.41 \\
            Baseline w/ one-hot context  & 49/54 & 0.91 & 42/54 & 0.78 \\
            Tactile context, vision + tactile & 48/54 & 0.89 & 38/54 & 0.70 \\
            \textbf{Text context, vision + tactile} &
            \textbf{51/54} & \textbf{0.94} & \textbf{45/54} & \textbf{0.83} \\
            Oracle w/ GT text context & 54/54 & 1.00 & 47/54 & 0.87 \\
            \midrule[1pt]
            \multicolumn{5}{l}{\textbf{Unseen objects} (6 objects)} \\
            \midrule
            Baseline w/o context & 14/54 & 0.26 & 8/54 & 0.15 \\
            Baseline w/ one-hot context  & 43/54 & 0.80 & 26/54 & 0.48 \\
            Tactile context, vision + tactile & 48/54 & 0.89 & 36/54 & 0.67 \\
            \textbf{Text context, vision + tactile} &
            \textbf{50/54} & \textbf{0.93} & \textbf{36/54} & \textbf{0.67} \\
            Oracle w/ GT text context & 53/54 & 0.98 & 36/54 & 0.67 \\
            \bottomrule
        \end{tabular}
    \end{subtable}
    \hfill
    \begin{subtable}[t]{0.49\textwidth}
        \centering
        \caption{Multi-object rearrangement task (three-bottle subset).}
        \label{tab:policy-results-set2}
        \scriptsize
        \setlength{\tabcolsep}{3.2pt}
        \renewcommand{\arraystretch}{1.10}
        \begin{tabular}{lcccc}
            \toprule
            \multirow{2}{*}{\textbf{Method}} &
            \multicolumn{2}{c}{\textbf{Strategy}} &
            \multicolumn{2}{c}{\textbf{Overall}} \\
            \cmidrule(lr){2-3}
            \cmidrule(lr){4-5}
            & Count & Rate & Count & Rate \\
            \midrule
            \multicolumn{5}{l}{\textbf{Seen objects} (3 objects)} \\
            \midrule
            Baseline w/o context & 28/45 & 0.62 & 25/45 & 0.56 \\
            Baseline w/o obj pos & 34/45 & 0.76 & 32/45 & 0.71 \\
            Tactile context, vision only & 37/45 & 0.82 & 33/45 & 0.73 \\
            Tactile context, vision + tactile & 37/45 & 0.82 & 36/45 & 0.80 \\
            Text context, vision only & 41/45 & 0.91 & 40/45 & 0.89 \\
            \textbf{Text context, vision + tactile} &
            \textbf{45/45} & \textbf{1.00} & \textbf{45/45} & \textbf{1.00} \\
            Oracle w/ GT text context & 45/45 & 1.00 & 43/45 & 0.96 \\
            \bottomrule
        \end{tabular}
    \end{subtable}

    \vspace{0.8em}
    \begin{subtable}[t]{0.49\textwidth}
        \centering
        \caption{Pouring task.}
        \label{tab:pouring-results}
        \scriptsize
        \setlength{\tabcolsep}{4pt}
        \renewcommand{\arraystretch}{1.10}
        \begin{tabular}{lcccc}
            \toprule
            \multirow{2}{*}{\textbf{Method}} &
            \multicolumn{2}{c}{\textbf{Strategy}} &
            \multicolumn{2}{c}{\textbf{Overall}} \\
            \cmidrule(lr){2-3}
            \cmidrule(lr){4-5}
            & Count & Rate & Count & Rate \\
            \midrule
            \multicolumn{5}{l}{\textbf{Seen objects} (2 objects)} \\
            \midrule
            Baseline w/o context & 10/20 & 0.50 & 10/20 & 0.50 \\
            Text context, vision only & 19/20 & 0.95 & 16/20 & 0.80 \\
            \textbf{Text context, vision + tactile} & \textbf{19/20} & \textbf{0.95} & \textbf{16/20} & \textbf{0.80} \\    
            \midrule
            \multicolumn{5}{l}{\textbf{Unseen objects} (1 object)} \\
            \midrule
            Baseline w/o context & 5/10 & 0.50 & 4/10 & 0.40 \\
            Text context, vision only & 9/10 & 0.90 & 8/10 & 0.80 \\
            \textbf{Text context, vision + tactile} & \textbf{9/10} & \textbf{0.90} & \textbf{7/10} & \textbf{0.70} \\
            \bottomrule
        \end{tabular}
    \end{subtable}
    \hfill
    \begin{subtable}[t]{0.49\textwidth}
        \centering
        \caption{Box opening task (with DeltaHand).}
        \label{tab:box-opening-results}
        \scriptsize
        \setlength{\tabcolsep}{4pt}
        \renewcommand{\arraystretch}{1.10}
        \begin{tabular}{lcccc}
            \toprule
            \multirow{2}{*}{\textbf{Method}} &
            \multicolumn{2}{c}{\textbf{Strategy}} &
            \multicolumn{2}{c}{\textbf{Overall}} \\
            \cmidrule(lr){2-3}
            \cmidrule(lr){4-5}
            & Count & Rate & Count & Rate \\
            \midrule
            \multicolumn{5}{l}{\textbf{Seen objects} (1 object)} \\
            \midrule
            Baseline w/o context & 12/48 & 0.25 & 12/48 & 0.25 \\
            \textbf{Text context, vision + tactile} & \textbf{48/48} & \textbf{1.00} & \textbf{48/48} & \textbf{1.00} \\
            \bottomrule
        \end{tabular}
    \end{subtable}
\end{table*}

\subsection{Manipulation Task Details}
\label{sec:task-appendix}
We evaluate four manipulation settings that require reasoning about different hidden object properties. Table~\ref{tab:task-required-property} summarizes the required properties for strategy selection in each task.
\begin{table}[h]
    \centering
    \caption{Manipulation tasks and the properties required for strategy selection.}
    \label{tab:task-required-property}
    \begin{tabular}{ll}
        \toprule
        Task & Required hidden property\\
        \midrule
        Multi-object rearrangement & rigidity + content  \\
        Three-container subset & rigidity + content  \\
        Pouring & fill level \\
        Box opening & weight / fill level\\
        \bottomrule
    \end{tabular}
\end{table}

\subsubsection{Multi-Object Rearrangement Task}
This task evaluates whether semantic object context generalizes across unseen object instances and hidden-state combinations. Given three objects placed simultaneously on a table, the policy must select appropriate manipulation strategies for each object during rearrangement.

\paragraph{Task Setup.}
The task consists of two steps: (1) \emph{Grasp selection}: empty objects are grasped from the top. For non-empty objects, the robot selects between top and side grasps based on object rigidity. 
Deformable objects require side grasps for stable manipulation, while rigid objects can be grasped from the top. Due to variations in object geometry and size, successful top grasps require accurate alignment with the object center before lowering the gripper. For side grasps, the robot must perform precise in-hand adjustments during approach to avoid unintended object displacement before grasp closure. (2) \emph{Placement decision}: objects are placed into target regions according to their internal contents. Objects containing liquid, granular, and solid contents are rearranged into the corresponding yellow, purple, and green shelf slots, while empty objects are placed into the recycle bin.

\paragraph{Generalization setup.}
The policy is trained on six objects and evaluated on six seen and six unseen objects, as shown in Fig.~\ref{fig:policy-results}(a). Since the same object can contain different internal contents, visually identical objects may require different manipulation strategies. 

The unseen scenarios can be a combination of the following two settings: 1) \textbf{Unseen objects}: objects with novel geometries and appearances; 2) \textbf{Unseen internal contents or fill level}: previously seen objects evaluated with unseen contents or fill levels. For example, a coffee container may appear empty or granular-filled during training but be evaluated as liquid-filled during testing.

\paragraph{Results and Failure Cases.}
As shown in Table~\ref{tab:policy-results-all}(a), object context is critical for multi-object rearrangement. 
For seen objects, the baseline policy produces 32/54 decision successes and 22/54 execution successes, whereas the text-context policy improves decision success to 51/54 with 45/54 execution successes. The improvement is more pronounced for unseen objects: the baseline achieves only 14/54 decision successes and 8/54 overall successes, while the text-context policy achieves 50/54 decision successes and 36/54 overall successes.

For the baseline policy, the dominant failures are incorrect decisions. Without object context, the policy cannot reliably infer rigidity or internal contents from visual observations alone, leading to incorrect grasp selection and placement targets. By contrast, context-conditioned policies substantially reduce decision failures. The remaining errors are mainly execution failures, including unstable side grasps of deformable containers, object slip during transport, and inaccurate placement into shelf slots. These failures occur more frequently for unseen objects, whose geometry and contact modes differ from those observed during training.

Comparison with oracle text context suggests that most remaining failures are not caused by semantic misclassification. On unseen objects, the predicted text-context policy and oracle policy achieve the same overall success rate (36/54), while their decision success rates are similarly high (50/54 vs.\ 53/54). The remaining failures are primarily due to low-level execution difficulties on novel object geometries.

\subsubsection{Multi-Object Rearrangement Task - Visually Ambiguous Three-Container Subset}

\paragraph{Task Setup.}
This task uses a subset of the previous rearrangement task consisting of three visually identical coffee containers (Fig.~\ref{fig:policy-results}(b)). It is designed to evaluate the role of semantic object context under severe visual ambiguity, and shares the same setup as the previous rearrangement task.

\paragraph{Results and Failure Cases.}
As shown in Table~\ref{tab:policy-results-all}(b), policies without object context suffer primarily from decision-making failures, since visually identical containers do not reveal whether an object is rigid or deformable, empty or filled. As a result, the policy often selects incorrect grasp types or placement targets. Providing object context substantially reduces these failures. Directly conditioning on tactile embeddings improves decision success from 28/45 to 37/45. Using text-aligned semantic context further improves decision success to 41/45 with vision-only observations and 45/45 when online tactile observations are added. This suggests that semantic representations provide a more effective abstraction for downstream decision-making than raw tactile embeddings.

The remaining failures are mainly execution failures. In some cases, unstable grasps lead to object slip during lifting or transport, while inaccurate placement causes objects to fall from the shelf. Adding online vibrotactile observations further improves manipulation robustness by providing reactive contact feedback and improving grasp stability during execution. With online tactile observations, the overall success rate of the tactile-context policy improves from 33/45 to 36/45, while the text-context policy improves from 40/45 to 45/45.

\subsubsection{Pouring Task}
In this task, the policy selects different manipulation strategies based on the container fill level. Empty containers are grasped from the top and directly recycled, while filled containers are grasped from the side, poured into the sink, and then recycled. This task further demonstrates the use of task-agnostic object representations for downstream manipulation.

\paragraph{Task Setup.} A single container is placed on the table in either an empty or water-filled state. Two containers (the green/brown can and pink can in Fig.~\ref{fig:policy-results}(c)) are used for training and seen-object evaluation, while an additional unseen container (red bottle) is used for generalization evaluation.

The task consists of three stages: (1) \emph{Grasp selection}: the policy first determines whether to use a top grasp for directly recycling an empty container or a side grasp for pouring. For filled containers, successful pouring requires stable side grasps with appropriate contact locations and grasp orientations to maintain stability during tilting. (2) \emph{Pouring}: the robot moves the container toward the sink and tilts the gripper to empty the liquid. Successful pouring requires stable grasping, appropriate wrist rotation, and accurate motion termination after the liquid is emptied. (3) \emph{Recycling}: after pouring, the robot places the container into the recycling bin.

\paragraph{Results and Failure Cases.} As shown in Table~\ref{tab:policy-results-all}(c), semantic object context is critical for decision-making. Without object context, the policy achieves near chance-level strategy selection, since visual observations alone cannot reliably distinguish empty from filled containers. As a result, the policy often overfits to a single manipulation strategy. With text context, the decision-making accuracy improves to 0.95 for seen objects and 0.90 for unseen objects, and few failures occur during execution.

Qualitative inspection of evaluation rollouts reveals a trade-off between grasp refinement and pouring-angle adaptation. Compared with the vision-only policy, the vision+tactile policy produces more object-adaptive gripper width commands, suggesting that online vibrotactile observations improve contact-aware grasp refinement.

However, successful pouring also requires adapting wrist tilting angles to the grasp pose and container geometry. In several vision+tactile failure cases, the robot executes the nominal pouring motion but fails to reach a sufficient tilting angle, preventing liquid flow. By contrast, in several successful vision-only rollouts, the robot continues tilting after observing visible liquid flow, leading to more complete emptying before recycling.

These observations suggest that tactile feedback is beneficial for grasp refinement and contact stability, but does not directly indicate whether liquid is flowing or whether the container has been fully emptied. For pouring tasks, visual and proprioceptive cues about container pose, liquid flow, and wrist orientation remain important for successful execution.

\subsubsection{Box Opening Task} 
In this task, the robot must identify and open the heavy box among a set of visually identical boxes based on weight-related object context. The task demonstrates that the proposed context-aware policy learning extends to dexterous robotic hands and enables manipulation behaviors that require coordinated in-hand motions, such as sliding a box tray open.

\paragraph{Task Setup.}
Four visually identical boxes are placed on the table, with one box containing screws while the others are empty. The box order is randomized. The task consists of two stages: (1) \emph{Approaching}: the robot chooses to approach the heavy box by moving the robot arm. For the empty box, the robot will take no actions. (2) \emph{Slide-to-open}: once positioned above the target box, the robotic hand uses two opposing fingers to stabilize the outer box while the remaining two fingers slide and push the inner tray outward until fully opened.

\paragraph{Results.} As shown in Table~\ref{tab:policy-results-all}(d), the baseline policy collapses to always attempting the box-opening behavior, resulting in a 0.25 success rate for both decision-making and overall task execution. With object context, the policy reliably selects the correct target box and successfully performs the opening behavior.

\end{document}